\documentclass{bmvc2k}

\usepackage{booktabs}
\usepackage{multirow}
\usepackage{amsthm}
\usepackage[capitalise]{cleveref}
\usepackage{amssymb}
\usepackage{floatrow}

\newfloatcommand{capbtabbox}{table}[][\FBwidth]

\newtheorem{assumption}{Assumption}
\newtheorem{proposition}{Proposition}
\newtheorem{hypothesis}{Hypothesis}

\title{Multi-Method Ensemble \\ for Out-of-Distribution Detection}

\addauthor{Lucas Rakotoarivony}{lucas.rakotoarivony@thalesgroup.com}{1}

\addinstitution{
Thales, cortAIx Labs \\ 
Palaiseau, 91120, France
}

\runninghead{Lucas Rakotoarivony}{MME}

\begin{document}

\maketitle

\begin{abstract}
Detecting out-of-distribution (OOD) samples is essential for neural networks operating in open-world settings, particularly in safety-critical applications. Existing methods have improved OOD detection by leveraging two main techniques: feature truncation, which increases the separation between in-distribution (ID) and OOD samples, and scoring functions, which assign scores to distinguish between ID and OOD data. However, most approaches either focus on a single family of techniques or evaluate their effectiveness on a specific type of OOD dataset, overlooking the potential of combining multiple existing solutions. Motivated by this observation, we theoretically and empirically demonstrate that state-of-the-art feature truncation and scoring functions can be effectively combined. Moreover, we show that aggregating multiple scoring functions enhances robustness against various types of OOD samples. Based on these insights, we propose the Multi-Method Ensemble (MME) score, which unifies state-of-the-art OOD detectors into a single, more effective scoring function. Extensive experiments on both large-scale and small-scale benchmarks, covering near-OOD and far-OOD scenarios, show that MME significantly outperforms recent state-of-the-art methods across all benchmarks. Notably, using the BiT model, our method achieves an average FPR95 of 27.57\% on the challenging ImageNet-1K benchmark, improving performance by 6\% over the best existing baseline.
\end{abstract}

\section{Introduction}
\label{sec:intro}

With the growing deployment of neural networks in real-world applications, they may encounter samples outside their trained class set. The unpredictable nature of such out-of-distribution (OOD) instances can prevent the model from making confident and reliable predictions \cite{yang2022openood}. Therefore, OOD detection is essential for ensuring the safety of machine learning systems, particularly in high-stakes applications such as autonomous driving \cite{chitta2021neat, vojir2021road}, medical imaging \cite{schlegl2019f, nair2020exploring} and biosynthesis \cite{vamathevan2019applications}.

\begin{figure}[t]
  \centering
   \includegraphics[width=0.8\linewidth]{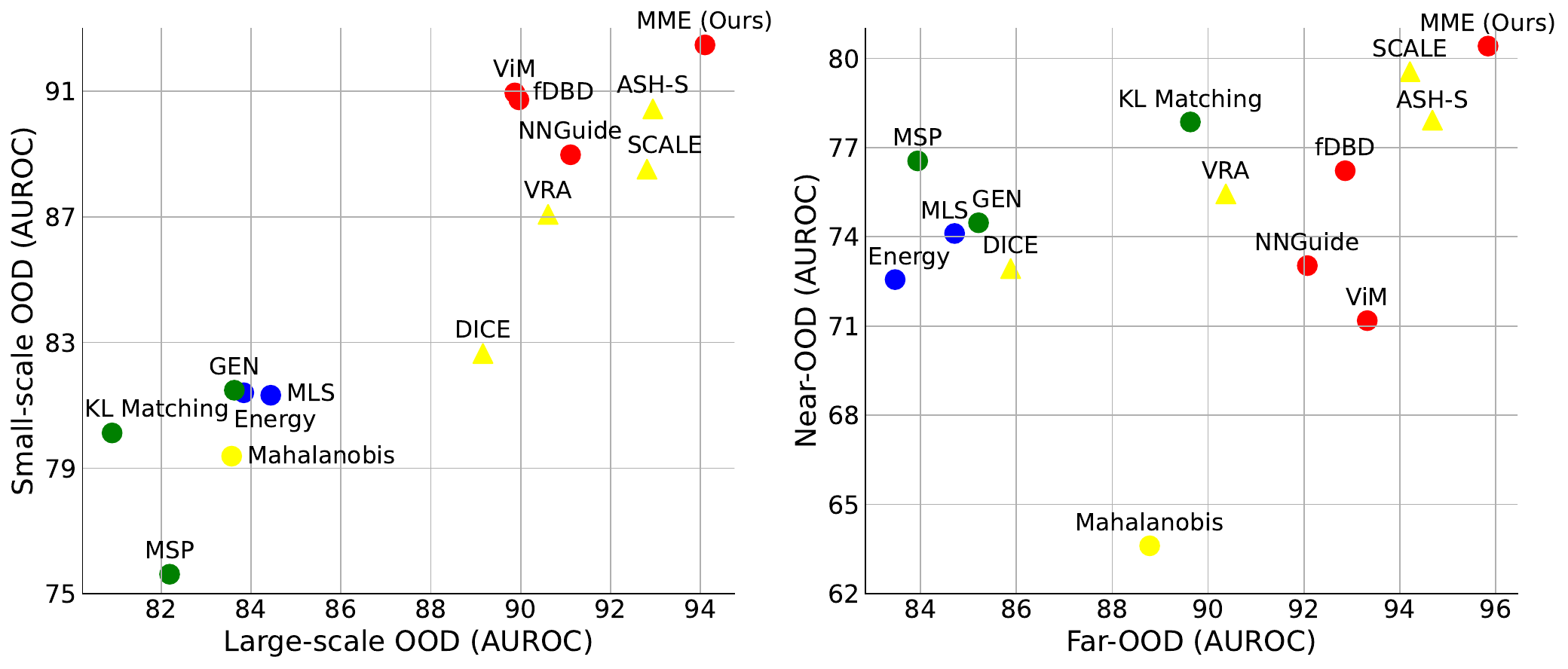}

   \caption{AUROC performance trade-offs for 14 algorithms, comparing small- vs. large-scale OOD detection (left) and near-OOD vs. far-OOD detection (right). Results are averaged across multiple datasets. For the small-scale setup, a DenseNet-100 model trained on CIFAR-100 is used, while the large-scale and near/far-OOD setups use a BiT model trained on ImageNet-1K. $\triangle$ denotes feature truncation methods combined with the Energy score, while $\bigcirc$ represents scoring functions. Green symbols indicate probability-based methods, blue symbols correspond to logit-based methods, yellow symbols represent feature-based methods, and red symbols denote approaches that leverage multiple sources of information.}
   \label{fig:near}
\end{figure}

The standard approach to OOD detection involves deriving a score function from a trained neural network, where in-distribution (ID) samples receive relatively higher scores than OOD samples. Researchers have developed scoring functions using either feature representations \cite{lee2018simple, sun2022out} or logit values \cite{hendrycks2019scaling, liu2020energy}, aiming to identify properties that naturally distinguish ID and OOD samples. More recent work \cite{wang2022vim, guan2023revisit, park2023nearest} has even combined both types of information sources to better handle the wide variety of OOD instances. However, while combining multiple inputs has been shown to be beneficial, these methods sometimes achieve suboptimal performance in specific applications \cite{zhang2023openood}, even compared to simpler scoring functions. In contrast, feature truncation methods \cite{sun2021react, sun2022dice} aim to increase the separation between ID and OOD samples by modifying the network’s propagated signals or weights. Despite their simplicity, feature truncation techniques have been shown to effectively reduce the overlap between ID and OOD samples at both the feature and logit levels.
However, these methods cannot be used independently and are typically combined with simple scoring functions, such as Energy \cite{liu2020energy} or MLS \cite{hendrycks2019scaling, vaze2021open}, for OOD detection. Therefore, even state-of-the-art feature truncation methods \cite{xu2023scaling, xu2023vra}, when paired with such scoring functions, often fail to outperform more advanced scoring methods due to the inherent simplicity of their approach.

In this paper, we empirically and theoretically demonstrate that combining multiple scoring functions with feature truncation methods improves OOD detection by better handling the diverse range of OOD examples encountered in real-world scenarios. Building on this ensemble approach, we propose a Multi-Method Ensemble (MME) score, which integrates recent scoring functions with feature truncation techniques. Specifically, the scoring process begins by extracting features, which are then refined using SCALE \cite{xu2023scaling} and VRA \cite{xu2023vra}
feature truncation to obtain sparse and clipped representations. These processed features are subsequently used by state-of-the-art scoring functions, including fDBD \cite{liu2023fast}, PCA \cite{guan2023revisit}, ViM \cite{wang2022vim}, and our proposed $\text{NME}^+$ method, to generate multiple scalar scores. Finally, these individual scores are combined to produce a final OOD score. The design of MME intuitively ensures that even if one method fails to detect an OOD sample, another method is likely to do so, thereby improving overall detection performance, as illustrated in \Cref{fig:near}.

To verify MME effectiveness, we perform experiments on several benchmark datasets and models, including CIFAR-10, CIFAR-100 \cite{krizhevsky2009learning}, and the more challenging ImageNet-1K \cite{russakovsky2015imagenet}. The results show that our method outperforms current post-hoc strategies, achieving state-of-the-art performance. The key contributions of this paper are summarized as follows:

\begin{itemize}
  \item \textbf{Theory:} We theoretically demonstrate that combining multiple scoring functions improves OOD detection. Additionally, we empirically show that scoring functions generally perform better when fused with feature truncation methods.
  \item \textbf{Methodology:} We propose an effective post-hoc strategy called MME, which combines multiple state-of-the-art OOD detection methods to better handle the wide variety of OOD samples.
  \item \textbf{Performance:} We evaluate MME on popular benchmarks, including both large-scale and small-scale scenarios, as well as near-OOD and far-OOD cases. Our results show that MME consistently outperforms existing post-hoc methods for OOD detection.
\end{itemize}

\section{Related Work}
\label{sec:related}

OOD detection research focuses on two approaches: feature truncation \cite{sun2021react} and creating a scalar score function to distinguish between OOD and ID samples \cite{hendrycks2016baseline}.

Feature truncation methods \cite{sun2021react, sun2022dice} aim to better separate ID and OOD samples by altering network signals or weights. ReAct \cite{sun2021react} clips high activation values to disrupt OOD signals while preserving ID ones. VRA \cite{xu2023vra} goes further by also suppressing low and amplifying mid-range activations. ASH \cite{djurisic2022extremely} prunes and scales activations, while SCALE \cite{xu2023scaling} avoids pruning to reduce ID-OOD overlap. However, feature truncation methods cannot be applied by themselves for detecting OOD and must be combined with a score function.

Scalar score functions for OOD detection are generally divided into classifier-based and distance-based approaches. Classifier-based methods, or confidence scores, use neural network outputs to derive scores. MSP \cite{hendrycks2016baseline} uses maximum softmax probability for ID detection, while MLS \cite{hendrycks2019scaling, vaze2021open} uses logits to preserve more information. Energy score \cite{liu2020energy} quantifies prediction uncertainty by mapping model outputs to an energy value.
Distance-based methods determine OOD samples by measuring distances in feature space. The Mahalanobis detector \cite{lee2018simple} uses class-wise mean distances and shared covariances. In contrast, fDBD \cite{liu2023fast} focuses on distances from decision boundaries, showing ID features lie further away.

Classifier-based detectors effectively identify near-OOD instances near decision boundaries but exhibit overconfidence in far-OOD regions \cite{hein2019relu}. In contrast, distance-based methods perform well in far-OOD detection but struggle with near-OOD samples close to ID classes. Recent methods \cite{wang2022vim, guan2023revisit, park2023nearest} address these limitations by combining classifier- and distance-based cues. ViM \cite{wang2022vim} combines virtual logits from feature residuals with an energy score, while NNGuide \cite{park2023nearest} improves logits-based confidence using KNN guidance in feature space.

\section{Methods}
\label{sec:methods}

\subsection{Preliminaries}
\label{subsec:preli}

To ensure consistency with previous work \cite{sun2021react, xu2023vra, park2023nearest}, we define the OOD detection task as follows: let \( \mathcal{X} \) be the input space and \( \mathcal{Y} \) be a label space containing \( k \) distinct categories. Consider a neural network \( f: \mathcal{X} \to \mathbb{R}^k \) trained on samples drawn from the in-distribution data. The goal of OOD detection is to design a scoring function, \( S \), that determines whether a test input \( \mathbf{x} \in \mathcal{X} \) belongs to the ID or OOD space based on its score \( S(\mathbf{x}) \) and a threshold $\tau$.

\begin{equation}
\mathbf{x} \in 
\begin{cases}
\text{ID} &  \text{if} \quad S(\mathbf{x}) \geq \tau \\
\text{OOD} & \text{if} \quad S(\mathbf{x}) < \tau
\end{cases}
\label{eq:definition}
\end{equation}

By convention, scoring functions are designed to produce high values for in-distribution data and low values for out-of-distribution data. Therefore, an effective scoring function should satisfy the following 

\begin{equation} 
\mathbb{E}_{\text{in}}[S(\mathbf{x})] \geq \mathbb{E}_{\text{out}}[S(\mathbf{x})] \label{eq:scoring}
\end{equation}

Let define $\mathbb{V}(S) = \mathbb{E}_{\text{in}}[S(\mathbf{x})] - \mathbb{E}_{\text{out}}[S(\mathbf{x})]$, by definition, OOD detection objective is to find $S$ that maximizes $\mathbb{V}(S)$.

Feature truncation aims to increase the separation between ID and OOD samples by adjusting the network's propagated signals or weights. Let $g$ be a feature truncation method and $z$ a feature vector, where $z = f(\mathbf{x})$. We adopt the terminology introduced by \cite{sun2021react, xu2023vra}.

\begin{equation}
\mathbb{E}_{\text{in}}[g(z)]-\mathbb{E}_{\text{out}}[g(z)] \geq \mathbb{E}_{\text{in}}[z]-\mathbb{E}_{\text{out}}[z] \label{eq:feature}
\end{equation}

For convenience, since scoring functions typically use feature vectors or logits to compute a confidence score, we adopt the notation $S(\mathbf{x})$ or $S(z)$, depending on the context.

In the \cref{ssec:theory}, we will demonstrate that combining two scoring functions can improve OOD detection performance. To achieve this, we make the following assumption:

\begin{assumption}
\label{ass:cov}
\textit{Let $S_1$ and $S_2$ be two scoring functions and $\text{Cov}(S_1,S_2)$ the covariance of $S_1$ and $S_2$. We assume $\text{Cov}_{\text{in}}(S_1,S_2) \geq \text{Cov}_{\text{out}}(S_1,S_2)$}
\end{assumption}

By definition, covariance is a measure of the degree to which two variables change together. Assumption \ref{ass:cov} means that two confidence scores will more likely produce correlated values when encountering ID data rather than OOD samples. 
To validate Assumption \ref{ass:cov}, we will use two classifiers that can also be employed for OOD detection: the well-known MLS \cite{hendrycks2016baseline} classifier (classifier-based) and the  Nearest-Mean-of-Exemplars (NME) classifier (distance-based) proposed by \cite{rebuffi2017icarl}. To ensure consistency with \Cref{eq:definition}, we will adapt the NME classifier to $\text{NME}^+$, defined as follows $\text{NME}^+(\mathbf{x}) =  \max(\frac{1}{\text{softmax}(\frac{\text{NME}(\mathbf{x})}{T})})$, where $T$ is a temperature parameter that controls softmax smoothness. For more details about $\text{NME}^+$, please refer to \cref{appendix:nme}.
Since $\text{NME}^+$ and MLS are two scoring functions that can be easily derived from classification tasks, we introduce a piecewise function called COnsistency (CO), along with its corresponding scoring function, $\text{CO}^+$, as follows:

{\small
\noindent\begin{minipage}{.5\linewidth}
\begin{align}
        \text{CO}(\mathbf{x})=\begin{cases}
        	1, & \text{if } c_{\text{NME+}}(\mathbf{x}) = c_{\text{MLS}}(\mathbf{x}), \\
            0, & \text{otherwise.}
    	\end{cases}. \label{eq:co}
\end{align}
\end{minipage}%
\begin{minipage}{.5\linewidth}
\begin{align}
        \text{CO}^+(\mathbf{x})=\begin{cases}
        	\lambda, & \text{if } c_{\text{NME+}}(\mathbf{x}) = c_{\text{MLS}}(\mathbf{x}), \\
            1, & \text{otherwise.}
    	\end{cases}.
\end{align}
\end{minipage}
}

\begin{figure}
\begin{floatrow}
\capbtabbox{%
  \begin{tabular}{cccc}
\toprule
& Dataset & ViT & BiT \\
\midrule
ID & ImageNet     & 0.90 & 0.74 \\ \midrule
\multirow{4}{*}{OOD} & iNaturalist & 0.43 & 0.32 \\
& Places      & 0.66 & 0.52 \\
& SUN         & 0.65 & 0.50 \\
& Textures    & 0.43 & 0.38 \\
\bottomrule
\end{tabular}
}{%
\caption{Consistency ratio evaluated on the ImageNet-1K dataset (ID) and four OOD datasets. Results are reported for both ViT and BiT architectures.}
\label{tab:co}
}
\ffigbox{%
  \includegraphics[width=0.9\linewidth]{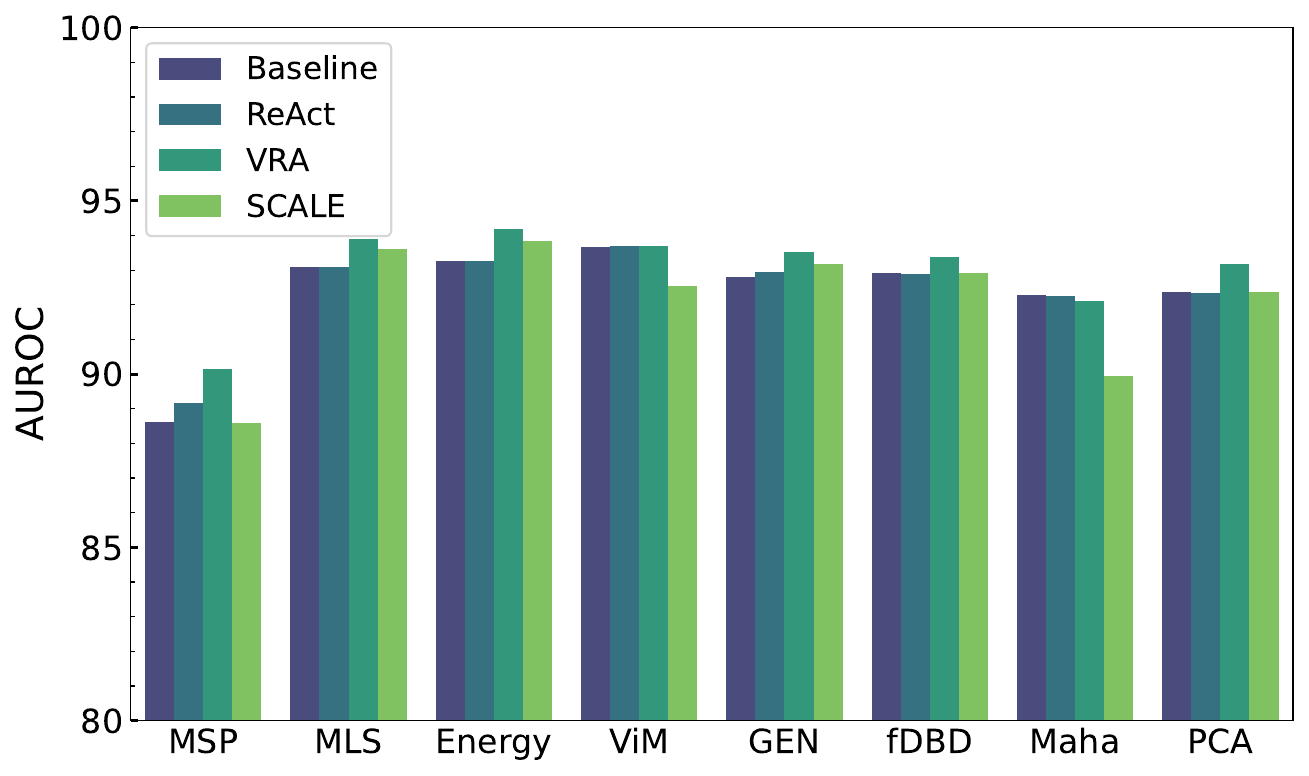}

}{%
  \caption{AUROC OOD detection performance on the ImageNet-1K benchmark for eight scoring functions combined with different feature truncation methods. Results are averaged across four OOD datasets.}
     \label{fig:truncation}

}
\end{floatrow}
\end{figure}

Intuitively, CO can be interpreted as a proxy for the covariance between $\text{NME}^+$ and MLS, as it measures agreement between their class predictions (denoted by $c$), while $\text{CO}^+$ is a scoring function derived from CO.
In $\text{CO}^+$, the parameter $\lambda \geq 1$ controls the influence of the prediction consistency between logits and features on the final score. \cref{tab:co} presents the consistency ratio for ID and OOD datasets across various models. From these results, we observe that $\text{NME}^+$ and MLS are more likely to make similar predictions when encountering ID data, compared to OOD data, which empirically validates Assumption \ref{ass:cov}. For simplicity, we use $\text{NME}^+$ and MLS as confidence scores in the definition of CO. However, as demonstrated in \cref{appendix:co}, the same reasoning applies equally well to other confidence scores.

\subsection{Motivation}

Current post-hoc OOD detection methods can be divided into two main categories: feature truncation \cite{sun2021react, sun2022dice} and scoring functions \cite{hendrycks2016baseline, liu2020energy}. Feature truncation methods aim to increase the gap between ID and OOD samples through scaling, pruning or clipping operations on features. However, feature truncation alone cannot perform OOD detection and is generally paired with simple scoring functions such as MLS \cite{hendrycks2016baseline} or Energy \cite{liu2020energy}. Scoring functions use various types of inputs, such as logits or features, to produce a score that can distinguish between ID and OOD samples. Recent advancements \cite{wang2022vim, guan2023revisit, park2023nearest} suggest using multiple input sources to better handle the diversity of OOD samples. These observations lead us to the following questions:

\begin{enumerate}
\item Does combining multiple scoring functions improve OOD detection?
\item Can feature truncation methods improve the performance of recent scoring functions?
\end{enumerate}

In this paper, we answer these questions through both theoretical and empirical analysis.

\subsection{Theoretical analysis} \label{ssec:theory}

\textbf{Combining multiple scoring function.} 
Recent state-of-the-art OOD detection methods have explored the benefits of combining feature and logits information \cite{wang2022vim, guan2023revisit} to better handle the diversity of OOD samples. To effectively merge these two sources of input, these methods typically combine the Energy score (logits-based information) with a custom scoring function based on features to achieve state-of-the-art performance. Building on these insights, we make the following proposition:

\begin{proposition}
\label{prop:scoring}
\textit{Let \( n \in \mathbb{N} \) with \( n \geq 2 \), and let \( S_1, \ldots, S_n \) be distinct scoring functions. Under Assumption \ref{ass:cov} and using \Cref{eq:scoring}, The inequality below is satisfied:
\[
\mathbb{V}(S_1 \cdot \ldots \cdot S_n) \geq \max\left( \mathbb{V}(S_1), \ldots, \mathbb{V}(S_n) \right).
\]}
\end{proposition}

\textit{Simplified proof.} The full proof is in \cref{appendix:proof}. For simplicity, we consider the case where $n = 2$, as the case $n> 2$ can be trivially derived by recurrence. Using the mathematical expression of the covariance and the scalability properties of a scoring function, we obtain:

\begin{align}
\begin{split}
\mathbb{V}(S_1 \cdot S_2) & \geq \mathbb{E}_{\text{in}}(S_1(\mathbf{x}))\mathbb{E}_{\text{in}}(S_2(\mathbf{x})) - \mathbb{E}_{\text{out}}(S_1(\mathbf{x}))\mathbb{E}_{\text{out}}(S_2(\mathbf{x}))\\
                        & \geq \mathbb{V}(S_1)
\end{split}
\end{align}

By symmetry, we have $\mathbb{V}(S_1 \cdot S_2) \geq \mathbb{V}(S_2)$, which leads to $\mathbb{V}(S_1 \cdot S_2) \geq \max(\mathbb{V}(S_1), \mathbb{V}(S_2))$

\cref{prop:scoring} states that combining multiple scoring functions results in a scoring function that separates ID and OOD data at least as effectively as the best individual scoring function, thus answering our first question. \\

\textbf{Pairing feature truncation with scoring function.} The second question is partially addressed in the literature. For instance, VRA \cite{xu2023vra} demonstrates that feature truncation increases the gap between ID and OOD distributions, as shown in \Cref{eq:feature}. Additionally, ReAct \cite{sun2021react} shows that clipping high activation values reduces both feature and logits activations more for OOD samples than for ID samples, leading to a greater separation between ID and OOD. Feature truncation methods \cite{sun2021react, xu2023vra, sun2022dice, xu2023scaling} indicate that this increased separation also transfers to the logits space and energy-based scores when combined with simple scoring functions, such as the Energy score \cite{liu2020energy} or MLS \cite{hendrycks2016baseline}. While recent scoring functions are more complex than Energy or MLS, they all rely on either features or logits as inputs. Based on these observations, we make the following hypothesis:

\begin{hypothesis}
\label{hypo:combine}
\textit{Let g be a feature truncation method such that $\mathbb{E}_{\text{in}}[g(z)]-\mathbb{E}_{\text{out}}[g(z)] \geq \mathbb{E}_{\text{in}}[z]-\mathbb{E}_{\text{out}}[z]$ and let $S$ be a scoring function. The following inequality holds:
\[
\mathbb{E}_{\text{in}}[S(g(z))]-\mathbb{E}_{\text{out}}[S(g(z))] \geq \mathbb{E}_{\text{in}}[S(z)]-\mathbb{E}_{\text{out}}[S(z)]
\]}
\end{hypothesis}

While \cref{hypo:combine} is central to our method, a general proof is elusive due to the nature of the scoring function $S$, which typically relies on the output of a neural network. Consequently, $S$ often does not satisfy properties such as monotonicity, Lipschitz continuity, or convexity. These conditions are commonly required to apply standard analytical tools like Jensen’s inequality or continuity bounds. As a result, we treat this statement as a working hypothesis. We support it empirically in \Cref{fig:truncation}, which shows that recent state-of-the-art scoring functions consistently benefit from feature truncation methods. It is important to note that Mahalanobis \cite{lee2018simple} is the only tested scoring function that does not benefit from feature truncation. This can be explained by the fact that, as a distance-based method, Mahalanobis relies on accurate feature distance values, which may be degraded by feature truncation.

\subsection{MME implementation}

Based on \cref{prop:scoring} and \cref{hypo:combine}, we have shown that various feature truncation methods and scoring functions can be combined to create a single scoring function maximizing the gap between ID and OOD. However, although it is theoretically possible to combine an infinite number of scoring functions, this approach is impractical due to computational overhead and does not necessarily guarantee better performance. Therefore, we focus on recent state-of-the-art feature truncation methods, namely SCALE \cite{xu2023scaling} and VRA \cite{xu2023vra}, along with scoring functions such as fDBD \cite{liu2023fast}, PCA \cite{guan2023revisit} and ViM \cite{wang2022vim}, as well as our proposed $\text{NME}^+$ and $\text{CO}^+$ functions, to construct MME scoring function. The MME function is defined as follows:

{\footnotesize
\begin{equation}
    \text{MME}(\mathbf{x}) = \exp({\text{SCALE}(\mathbf{x}) - \text{ViM}(\text{VRA}(\mathbf{x}))}) \cdot \text{fDBD}(\text{VRA}(\mathbf{x})) \cdot \text{PCA}(\text{VRA}(\mathbf{x})) \cdot \text{CO}^+(\mathbf{x}) \cdot \text{NME}^+(\mathbf{x}) 
\end{equation}
}

\begin{figure*}[t]
  \centering
   \includegraphics[width=0.8\linewidth]{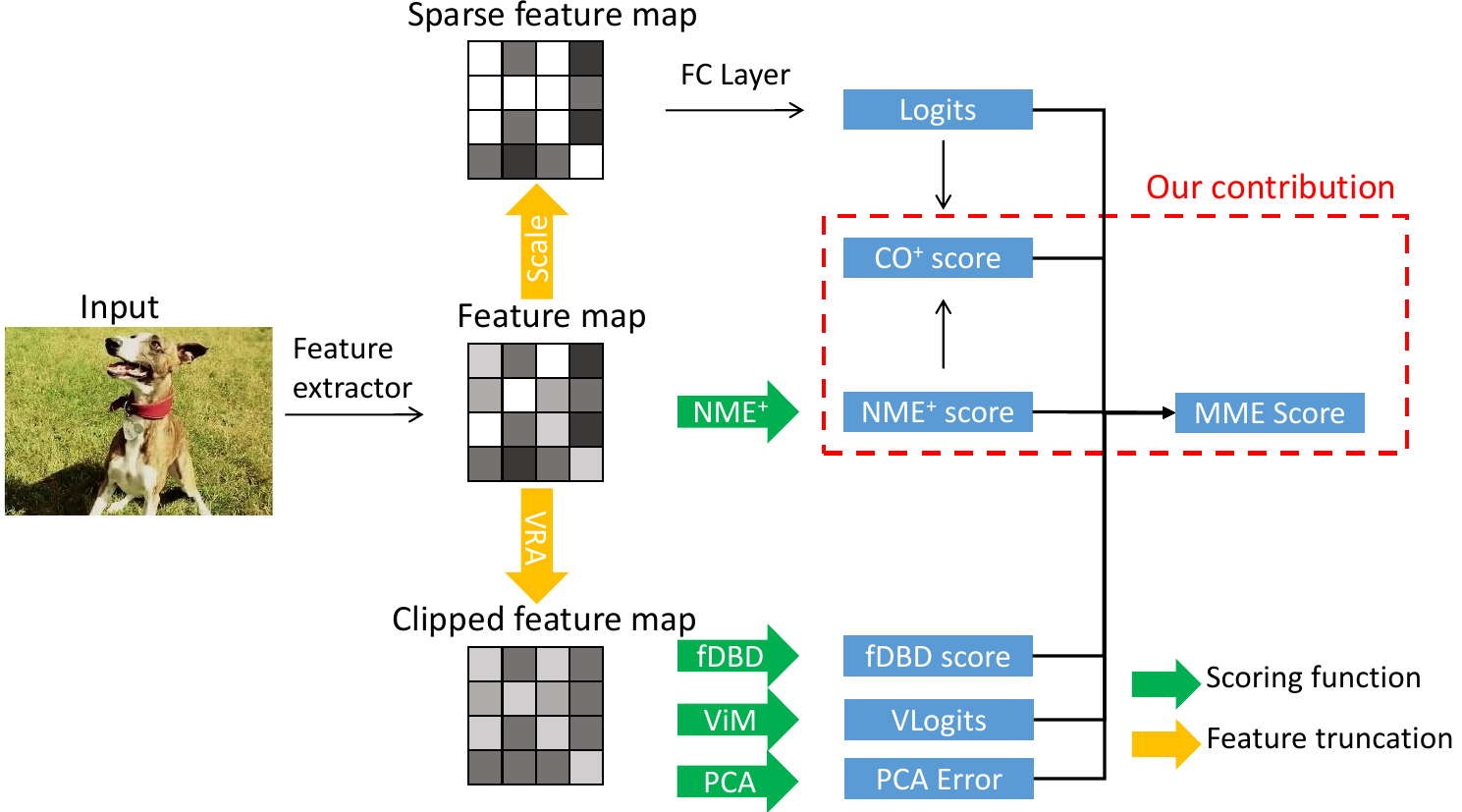}

   \caption{Overview of the MME pipeline. First, features are extracted from the input and processed using the VRA and SCALE feature truncation methods. Next, multiple scoring functions are applied to the feature maps to generate scalar scores. Finally, these scores are combined to produce MME score, which can be used with a threshold $\tau$ for OOD detection.}
   \label{fig:overview}
\end{figure*}

To ensure positivity of MME, we apply the exponential function to ${\scriptsize \text{SCALE}(\mathbf{x}) - \text{ViM}(\text{VRA}(\mathbf{x}))}$. Note that $\text{NME}^+$ is the only scoring function that does not use feature truncation, as it is a distance-based method, as suggested by \Cref{fig:truncation}. Further details about the choice of combinations will be provided in \cref{appendix:ablation}. A complete overview of MME pipeline is provided by \Cref{fig:overview}.

\section{Experiments}
\label{sec:experiments}

\subsection{Experimental settings}

\textbf{Datasets} MME was evaluated on three OOD detection benchmarks, each with training ID, test ID, and multiple OOD sets that don't class overlap with the ID sets. For small-scale benchmarks, CIFAR-10 and CIFAR-100 \cite{krizhevsky2009learning} serve as ID sets, with six OOD sets: LSUN-Crop \cite{yu2015lsun}, LSUN-Resize \cite{yu2015lsun}, iSUN \cite{xu2015turkergaze}, SVHN \cite{netzer2011reading}, Places365 \cite{zhou2017places}, and Textures \cite{cimpoi2014describing}. For large-scale OOD detection, ImageNet-1K \cite{russakovsky2015imagenet} is the ID set, with iNaturalist \cite{van2018inaturalist}, Places \cite{zhou2017places}, SUN \cite{xiao2010sun}, and Textures \cite{cimpoi2014describing} as OOD sets. Following recent studies \cite{xu2023scaling, zhang2023openood}, ImageNet-1K \cite{russakovsky2015imagenet} is used for near-OOD (NINCO \cite{bitterwolf2023or}, SSB-hard \cite{vaze2021open}) and far-OOD (iNaturalist \cite{van2018inaturalist}, Textures \cite{cimpoi2014describing}, OpenImage-O \cite{wang2022vim}) detection. Refer to \cref{appendix:dataset} for dataset details.

\textbf{Models} Following previous studies \cite{wang2022vim, liu2023gen, liang2017enhancing}, we used ViT-B/16 \cite{dosovitskiy2020image} (Transformer-based) and BiT-S-R101x1 \cite{kolesnikov2020big} (CNN-based) as the classifier backbones for the ImageNet-1K benchmarks, while DenseNet-BC-100 \cite{huang2017densely} was used for the CIFAR benchmarks.

\textbf{Methods} To ensure a fair comparison, we compare MME with several state-of-the-art post-hoc OOD detectors. We consider 5 feature truncation methods combined with the Energy score: ReAct \cite{sun2021react}, DICE \cite{sun2022dice}, ASH-S \cite{djurisic2022extremely}, SCALE \cite{xu2023scaling} and VRA \cite{xu2023vra}. We also evaluate 11 scoring functions: MSP \cite{hendrycks2016baseline}, MLS \cite{hendrycks2019scaling, vaze2021open}, Energy \cite{liu2020energy}, Mahalanobis \cite{lee2018simple} (Maha), KL Matching \cite{hendrycks2019scaling} (KL), ViM \cite{wang2022vim}, SHE \cite{zhang2022out}, GEN \cite{liu2023gen}, PCA \cite{guan2023revisit}, NNGuide \cite{park2023nearest}, and fDBD \cite{liu2023fast}. We use the same hyperparameters as in the original papers and reproduce results using their code. Experiments were conducted on an NVIDIA GeForce RTX 3090, with CUDA 11.0 and PyTorch 1.7. In the following experiments, $T=0.5$ for the ImageNet-1k benchmark and $T=0.1$ for the CIFAR benchmark, while $\lambda$ is set to 2 for both benchmarks. MME is robust to $T$ and $\lambda$, requiring minimal hyperparameter tuning, as shown in \cref{ssec:hyperpara}.

\textbf{Evaluation metrics} We report two metrics: AUROC, the area under the ROC curve (higher is better), and FPR95 (false positive rate at 95\% true positive rate, lower is better). Both metrics are presented as percentages, with the best results in bold.

\subsection{Evaluation on ImageNet-1k}
\label{subsec:imagenet}

On the ImageNet-1K large-scale benchmark, \cref{tab:imagenet} summarizes the OOD detection performance of each method across different OOD datasets, as well as the average performance over all four datasets. With the BiT backbone, MME outperforms all methods on three out of the four datasets in terms of FPR95, achieving an average FPR95 of 27.57\%, an improvement of 6\% over the best existing baseline. Furthermore, with the ViT model, MME achieves the highest performance on the Textures dataset and the best average performance overall. On the other three datasets, MME consistently delivers competitive results, generally ranking among the top methods (second or third), further highlighting its robustness.

Across all experiments, at least nine different methods achieve a top-three performance on a specific model and dataset, highlighting the sensitivity of existing OOD detectors to both the model architecture and dataset characteristics. In contrast, MME consistently ranks among the top three, demonstrating that our ensemble strategy enhances robustness and improves the detection of various types of OOD samples.

\subsection{Evaluation on CIFAR benchmarks}

On small-scale CIFAR benchmarks, the ensembling strategy also contributes to achieving state-of-the-art performance with greater consistency. \cref{tab:cifar} presents the average performance across six OOD datasets for both CIFAR-10 and CIFAR-100 as ID sets, demonstrating that MME outperforms existing post-hoc detectors. Notably, while ViM achieves the second-best performance on both CIFAR-10 and CIFAR-100 benchmarks and even comes within 1\% of MME in terms of FPR95 on CIFAR-10, it falls significantly short on CIFAR-100. In this case, MME achieves an FPR95 of 28.96\%, offering a substantial improvement of nearly 10\% over ViM, which records an FPR95 of 38.36\%.

\begin{table*}[b]
\scriptsize
\begin{tabular}{ccccccccccc}
\toprule
\multirow{2}{*}{ID Dataset} & \multirow{2}{*}{Metrics} & \multicolumn{8}{c}{Methods} \\ \cmidrule(l){3-11}
      &                     & \footnotesize{MSP}   & \footnotesize{MLS} & \footnotesize{ViM} & \footnotesize{ASH-S} & \footnotesize{GEN} & \footnotesize{SCALE} & \footnotesize{VRA} & \footnotesize{fDBD} & \footnotesize{MME} \\ \midrule
\multirow{2}{*}{CIFAR-10}           & \multicolumn{1}{|l|}{AUROC$\uparrow$}                         & 92.17  & 93.48   & 96.58 & 94.93 & 94.88 & 93.90 & 94.92 & 96.14 & \textbf{96.92}    \\
           &  \multicolumn{1}{|l|}{FPR95$\downarrow$}                   &  47.11 & 29.63  & 16.48 & 23.36  & 30.01 & 28.59 & 27.14 & 20.52 & \textbf{15.60} \\ \midrule
\multirow{2}{*}{CIFAR-100}           & \multicolumn{1}{|l|}{AUROC$\uparrow$}                         & 75.62  & 81.32  & 90.95 & 90.44 & 81.48 & 88.52 & 87.08 & 90.73 & \textbf{92.48}   \\
           &  \multicolumn{1}{|l|}{FPR95$\downarrow$}                   & 77.52  & 68.64  &   38.36 & 40.95 & 68.54 & 47.59 & 61.63 & 44.61 & \textbf{28.96}  \\  \bottomrule
\end{tabular}
\caption{Comparison between different methods in OOD detection on small-scale CIFAR benchmarks using a DenseNet model. Values are average percentages over six OOD datasets.}
\label{tab:cifar}
\end{table*}

\subsection{Evaluation on near- and far-OOD}

\Cref{fig:near} illustrates the trade-off between near-OOD and far-OOD detection performance across various state-of-the-art methods. As highlighted by \cite{park2023nearest}, simple probability-based methods such as MLS \cite{hendrycks2019scaling} and KL Matching \cite{hendrycks2019scaling} achieve competitive results on near-OOD but perform poorly on far-OOD. In contrast, distance-based scoring functions such as Mahalanobis \cite{lee2018simple} struggle with near-OOD detection due to the difficulty in defining clear decision boundaries. Recent methods like fDBD \cite{liu2023fast} and ViM \cite{wang2022vim}, which integrate multiple input sources, offer a better balance between near-OOD and far-OOD performance. Additionally, feature truncation techniques such as SCALE \cite{xu2023scaling} and ASH-S \cite{djurisic2022extremely} achieve a strong trade-off due to their effective scaling properties. Ultimately, MME outperforms all other approaches on both near-OOD and far-OOD tasks, further emphasizing the advantage of combining feature truncation with multiple scoring functions to detect a wide variety of OOD samples.

\begin{table*}[]
\scriptsize
\begin{tabular}{ccccccccccc}
\toprule
\multirow{2}{*}{Method}  & \multicolumn{2}{c}{iNaturalist} & \multicolumn{2}{c}{Places} & \multicolumn{2}{c}{SUN} & \multicolumn{2}{c}{Textures} & \multicolumn{2}{c}{Average} \\
                                          & \tiny{AUROC$\uparrow$}        & \tiny{FPR95$\downarrow$}         & \tiny{AUROC$\uparrow$}             & \tiny{FPR95$\downarrow$}             & \tiny{AUROC$\uparrow$}           & \tiny{FPR95$\downarrow$}            & \tiny{AUROC$\uparrow$}              & \tiny{FPR95$\downarrow$}               & \tiny{AUROC$\uparrow$}             & \tiny{FPR95$\downarrow$}              \\ \cmidrule(l){1-1} \cmidrule(l){2-9} \cmidrule(l){10-11}
                                          \multicolumn{11}{c}{Backbone : BiT} \\
                                          \cmidrule(l){1-1} \cmidrule(l){2-9} \cmidrule(l){10-11}
                                        MSP \tiny{(ICLR'17)} & 87.90            & 64.58               & 79.27              & 79.00            & 81.82            & 71.82           & 79.76              & 77.17             & 82.19             & 73.14 \\
                                         MLS \tiny{(ICML'22)} &   86.78             & 70.51               & 82.58              & 74.36            & 86.74          & 63.49            & 81.65              & 73.60             & 84.44             & 70.49 \\
                                         Energy \tiny{(NeurIPS'20)} & 84.52               & 74.96                 & 82.56              & 73.21           & 87.17            & 60.81           & 81.10              & 73.91             &  83.84            & 70.72 \\
                                         ReAct \tiny{(NeurIPS'21)} & 91.50               &  48.65              & 87.96             & 54.67            & 92.33          & 38.99            & 90.65              & 50.14             & 90.61             & 48.11 \\
                                         Maha \tiny{(NeurIPS'18)} & 85.80               & 64.69               & 73.06             & 82.35            &    78.10        & 73.38           &    97.33           & 13.95             & 83.57             & 58.59 \\ 
                                         KL \tiny{(ICML'22)} & 90.84               &    \underline{38.19}            & 73.17             & 79.10            & 76.43           & 75.88           & 83.22              & 55.16             &  80.91            & 62.08 \\
                                         ViM \tiny{(CVPR'22)} & 89.38              & 55.09                &  83.33            & 67.94            & 87.85           & 57.79           &  \textbf{98.92}             & \textbf{4.63}             & 89.87             & 46.36 \\
                                         DICE \tiny{(ECCV'22)} & 91.44               & 43.64               & \underline{88.99}             & \underline{49.20}            & \underline{92.52}           &    35.54        & 83.67              & 55.10             &  89.16            & 45.87 \\
                                         ASH-S \tiny{(ICLR'23)} & \underline{93.87}            & \underline{36.85}               & 88.49             & \underline{49.17}            &    91.77        & \underline{34.69}           &    \underline{97.62}           & \underline{12.03}             & \underline{92.94}             & \underline{33.19} \\
                                         SHE \tiny{(ICLR'23)} & 77.65               &    72.25            & 80.31             &  71.63           & 84.65           & 56.68           & 84.58              & 48.84             & 81.80             & 62.35 \\
                                         GEN \tiny{(CVPR'23)} & 88.67               &  68.32              &  80.19            & 79.66            & 84.17           & 72.23           & 81.48              & 77.91             & 83.63             & 74.53 \\
                                         PCA \tiny{(ICCV'23)} & 50.37               &    99.77            & 42.19             & 99.80            & 50.15           & 99.25           & 69.69              & 85.35             & 53.1              & 96.04 \\
                                         NNGuide \tiny{(ICCV'23)} & 88.48               &    52.83            & 88.17             & 50.76            & \textbf{92.92}           & \underline{33.58}           &  94.94             & 21.51             &  91.11            & 40.60 \\
                                         SCALE \tiny{(ICLR'24)} & \underline{93.77}               & 39.37               &    \underline{88.96}          &    50.10         & 92.39           &  35.28          & 96.11              &    18.99          &  \underline{92.81}            & \underline{35.94} \\
                                         VRA \tiny{(NeurIPS'24)} & 91.50                & 48.65               & 87.96              & 54.67             & 92.33          &    38.99         & 90.65             & 50.14              & 90.61             & 48.11 \\
                                         fDBD \tiny{(ICML'24)} & 91.71               & 51.42               &  84.20            & 67.45            & 88.47           &    56.03        & 95.45              & 24.38             &  89.96            & 49.82\\
                                         MME \tiny{(Ours)} & \textbf{95.53}               & \textbf{26.91}               &    \textbf{89.73}          &    \textbf{45.03}         & \underline{92.70}           & \textbf{30.64}           & \underline{98.45}             &    \underline{7.71}          & \textbf{94.10}             & \textbf{27.57}
                                         
                                         \\ \cmidrule(l){1-1} \cmidrule(l){2-9} \cmidrule(l){10-11}
                                          \multicolumn{11}{c}{Backbone : ViT} \\
                                          \cmidrule(l){1-1} \cmidrule(l){2-9} \cmidrule(l){10-11}
                                          MSP \tiny{(ICLR'17)} & 96.13               & 19.15               &    85.09          &   59.97          &  86.13          &  57.00          & 87.13               & 48.45             &  88.62            & 46.14 \\ 
                                         MLS \tiny{(ICML'22)} & 98.57               & 6.53                & 89.24             & 46.65            & 91.50            &  38.97          & 93.05               &  30.27            &    93.09          & 30.61 \\
                                         Energy \tiny{(NeurIPS'20)} & 98.66               &  6.04              &  89.30             &  \underline{45.27}           & 91.75           &  37.09          &  93.42             & 28.12              &  93.28            & 29.13 \\
                                         ReAct \tiny{(NeurIPS'21)} & 99.01               & 4.19              & 89.11             & 47.37            & 91.54          & 39.04            & 93.38               & 28.37             & 93.26             & 29.74\\
                                         Maha \tiny{(NeurIPS'18)} & \textbf{99.63}               & \underline{2.15}               & 86.29             & 60.04            & 89.11           & 51.29           & \underline{94.21}               & \underline{25.27}            & 92.31             & 34.69 \\
                                         KL \tiny{(ICML'22)} & 96.62               & 16.45               & 84.34             & 63.50            & 85.91            & 59.41           & 88.26               & 45.99          & 88.78             & 46.34 \\
                                         ViM \tiny{(CVPR'22)} & 99.41                & 2.56               & 88.47             &  50.44            & 91.53            & 39.48            & \underline{95.31}              & \underline{20.14}              &  93.68            & 28.15 \\
                                         DICE \tiny{(ECCV'22)} & 65.49              & 91.69               & 61.56              & 93.88            & 64.75           & 88.51           & 75.49              & 74.26             &  66.82            & 87.09 \\
                                         ASH-S \tiny{(ICLR'23)} & 61.96               &  95.34              & 61.23             &  95.68           & 65.52           & 92.90           & 56.79              & 91.12              & 61.38             & 93.76 \\
                                         SHE \tiny{(ICLR'23)} & 90.69               & 43.50               &         85.63     & 59.25            & 87.97           &  51.93          & 92.18              & 32.31             & 89.12             & 46.75\\
                                         GEN \tiny{(CVPR'23)} & 98.63               & 5.83               & 89.22              &  49.61           & 91.00           & 43.36           & 92.35              & 34.05             & 92.80             & 33.21 \\
                                         PCA \tiny{(ICCV'23)} & 95.36              & 27.37                & 76.36              & 79.58            & 80.92          & 72.92            & 90.20              & 43.59             & 85.71             & 55.86 \\
                                         NNGuide \tiny{(ICCV'23)} & 87.25               & 52.54               & 80.85            & 62.61             & 86.04            & 51.16           & 91.97              & 32.66             &  88.03            & 46.05 \\
                                         SCALE \tiny{(ICLR'24)} & 98.88                & 5.00               & \underline{90.14}             & \textbf{42.89}            & \underline{92.57}           & \textbf{34.28}           & 93.39             & 28.51               & \underline{93.75}             & \underline{27.67} \\
                                         VRA \tiny{(NeurIPS'24)} & \underline{99.52}                &  \textbf{1.89}              &  \textbf{90.75}            & \underline{42.98}            & \textbf{92.99}        &  \underline{34.49}          & 93.37              & 29.19              & \underline{94.16}              & \underline{27.14} \\
                                         fDBD \tiny{(ICML'24)} & 98.39               & 7.15               & 89.07             &  48.68             &  91.13           & 41.35           & 93.10              & 29.81             & 92.92             & 31.75\\
                                         MME \tiny{(Ours)} & \underline{99.48}               & \underline{2.16}                & \underline{89.66}             & 46.71            & \underline{92.49}          & \underline{35.92}             & \textbf{95.48}              & \textbf{19.77}             & \textbf{94.28}             & \textbf{26.14} \\
                                         \bottomrule
\end{tabular}
\caption{OOD detection for MME and state-of-the-art methods. The ID dataset is ImageNet-1K and the OOD datasets are iNaturalist, Places, SUN and Textures. Results are reported using AUROC and FPR95 metrics (in percentage). A pre-trained BiT and a ViT model are tested. The best method is highlighted in bold, and the 2nd and 3rd best are underlined.}
\label{tab:imagenet}
\end{table*}

\section{Conclusion}

In this study, we analyzed existing feature truncation and scoring function methods and demonstrated both theoretically and empirically that combining these methods improves OOD detection performance. Based on this insight, we proposed a MME strategy that fuses state-of-the-art OOD detectors to better handle the diversity of OOD samples. MME has been thoroughly tested on both large and small-scale benchmarks, considering both near- and far-OOD scenarios. MME has been shown to achieve state-of-the-art performance across various models and datasets, demonstrating its effectiveness and robustness in detecting a wide variety of OOD samples.

\bibliography{mme}

\newpage
\appendix

\section{Mathematical proof}
\label{appendix:proof}

Here, we provide the proof of \cref{prop:scoring}.

\textbf{Proposition 1} \textit{Let $n\in\mathbb{N}$ such that $n\geq2$. Let $S_1, \ldots, S_n$ different scoring functions. $\mathbb{V}(S_1 \cdot \ldots \cdot S_n) \geq \max(\mathbb{V}(S_1), \ldots, \mathbb{V}(S_n))$}

\textit{Proof.} We start with the case where $n=2$. By definition, we have : 

\begin{align}
\begin{split}
\mathbb{V}(S_1 \cdot S_2)&  =  \mathbb{E}_{\text{in}}[S_1 \cdot S_2(\mathbf{x})] - \mathbb{E}_{\text{out}}[S_1 \cdot S_2(\mathbf{x})] \\ 
& = \mathbb{E}_{\text{in}}(S_1(\mathbf{x}))\mathbb{E}_{\text{in}}(S_2(\mathbf{x})) + \text{Cov}_{\text{in}}(S_1,S_2) - \mathbb{E}_{\text{out}}(S_1(\mathbf{x}))\mathbb{E}_{\text{out}}(S_2(\mathbf{x})) - \text{Cov}_{\text{out}}(S_1,S_2)
\end{split}
\end{align}

However, thanks to Assumption \ref{ass:cov}, we have $\text{Cov}_{\text{in}}(S_1,S_2) \geq \text{Cov}_{\text{out}}(S_1,S_2)$ resulting in,

\begin{align}
\begin{split}
\mathbb{V}(S_1 \cdot S_2) \geq & \mathbb{E}_{\text{in}}(S_1(\mathbf{x}))\mathbb{E}_{\text{in}}(S_2(\mathbf{x})) - \mathbb{E}_{\text{out}}(S_1(\mathbf{x}))\mathbb{E}_{\text{out}}(S_2(\mathbf{x}))
\end{split}
\end{align}

By definition of a scoring function and \cref{eq:scoring}, we have $\mathbb{E}_{\text{in}}[S_2(\mathbf{x})] \geq \mathbb{E}_{\text{out}}[S_2(\mathbf{x})]$ resulting in,

\begin{align}
\begin{split}
\mathbb{V}(S_1 \cdot S_2) &\geq \mathbb{E}_{\text{in}}(S_1(\mathbf{x}))\mathbb{E}_{\text{out}}(S_2(\mathbf{x})) - \mathbb{E}_{\text{out}}(S_1(\mathbf{x}))\mathbb{E}_{\text{out}}(S_2(\mathbf{x})) \\
& \geq (\mathbb{E}_{\text{in}}(S_1(\mathbf{x}))-\mathbb{E}_{\text{out}}(S_1(\mathbf{x})))\mathbb{E}_{\text{out}}(S_2(\mathbf{x})) \\
& \geq \mathbb{V}(S_1)\mathbb{E}_{\text{out}}(S_2(\mathbf{x}))
\end{split}
\end{align}

By definition, a scoring function is invariant to scaling, therefore if $\mathbb{E}_{\text{out}}(S_2(\mathbf{x})) < 1$, we could easily find a $\lambda$ such as $\mathbb{E}_{\text{out}}(\lambda S_2(\mathbf{x}))\geq1$. For convenience and clarity, we will suppose $\mathbb{E}_{\text{out}}(S_2(\mathbf{x})) \geq 1$ ultimately leading us to

\begin{equation}
    \mathbb{V}(S_1 \cdot S_2) \geq \mathbb{V}(S_1) 
\end{equation}

The same methodology can be applied to obtain $\mathbb{V}(S_1 \cdot S_2) \geq \mathbb{V}(S_2)$, ultimately leading to $\mathbb{V}(S_1 \cdot S_2) \geq \max(\mathbb{V}(S_1), \mathbb{V}(S_2))$.

For the case $n > 2$, we can trivially obtain the result by recurrence by defining $S_1 = S_1 \cdot \ldots \cdot S_{n-1}$ and $S_2 = S_n$ which completes the proof.

\section{Detailed Description of $\text{NME}^+$ and CO}
\label{appendix:nme}

\subsection{Motivation of $\text{NME}^+$ and CO}

The first OOD detectors, MSP \cite{hendrycks2016baseline} and MLS \cite{hendrycks2019scaling, vaze2021open}, rely directly on the output of the classifier. The underlying assumption is that in-distribution samples are more likely to receive higher confidence scores than OOD samples, as the model is generally more "confident" in its predictions for familiar data. This principle allows us to easily construct an OOD detector based on any classifier following the same logic.

While MSP and MLS are classifier-based methods, we propose using the Nearest-Mean-of-Exemplars (NME) classifier, a distance-based approach originally introduced in \cite{rebuffi2017icarl} to better leverage feature representations. Furthermore, as discussed in \cref{sec:related}, classifier-based and distance-based OOD detectors are complementary. Therefore, combining both approaches through the CO function can enhance OOD detection performance across a broader range of scenarios and allows us to empirically validate Assumption \ref{ass:cov}.

\subsection{Implementation of $\text{NME}^+$}

Let $\phi$ be the feature extractor. For each class $y \in \mathcal{Y}$, the mean feature vector is computed as:
\[
\mu_y = \frac{1}{|\mathcal{P}_y|} \sum_{x \in \mathcal{P}_y} \phi(x)
\]

Given a test input $x \in \mathcal{X}$, its feature representation $\phi(x)$ is compared to all class means, and the predicted label is assigned as:
\[
\text{NME}(\mathbf{x}) = \arg\min_{y \in \mathcal{C}} \left\| \phi(\mathbf{x}) - \mu_y \right\|_2
\]

To ensure positivity and remain consistent with the definition of scoring function and \Cref{eq:definition}, we use $\text{NME}^+$ as a scoring function.

\[
\text{NME}^+(\mathbf{x}) =  \max(\frac{1}{\text{softmax}(\frac{\text{NME}(\mathbf{x})}{T})})
\]

\subsection{Performance of  $\text{NME}^+$ and $\text{CO}^+$}

\cref{tab:nme} presents the raw performance of $\text{NME}^+$ and $\text{CO}^+$ compared to the full MME framework and other state-of-the-art methods. Despite its relative simplicity, $\text{NME}^+$ achieves competitive results, comparable to Mahalanobis and even GEN. When combined with $\text{CO}^+$, it yields a 1.26\% improvement in average FPR95, highlighting the benefit of integrating MLS with NME-based predictions. Furthermore, as demonstrated in \cref{subsec:imagenet}, incorporating both $\text{NME}^+$ and $\text{CO}^+$ into other state-of-the-art approaches to form MME leads to superior performance and enhanced robustness across diverse OOD scenarios.

\begin{table*}[]
\scriptsize
\begin{tabular}{ccccccccccc}
\toprule
\multirow{2}{*}{Method}  & \multicolumn{2}{c}{iNaturalist} & \multicolumn{2}{c}{Places} & \multicolumn{2}{c}{SUN} & \multicolumn{2}{c}{Textures} & \multicolumn{2}{c}{Average} \\
                                          & \tiny{AUROC$\uparrow$}        & \tiny{FPR95$\downarrow$}         & \tiny{AUROC$\uparrow$}             & \tiny{FPR95$\downarrow$}             & \tiny{AUROC$\uparrow$}           & \tiny{FPR95$\downarrow$}            & \tiny{AUROC$\uparrow$}              & \tiny{FPR95$\downarrow$}               & \tiny{AUROC$\uparrow$}             & \tiny{FPR95$\downarrow$}              \\ \cmidrule(l){1-1} \cmidrule(l){2-9} \cmidrule(l){10-11}
                                          
                                          MSP \tiny{(ICLR'17)} & 96.13               & 19.15               &    85.09          &   59.97          &  86.13          &  57.00          & 87.13               & 48.45             &  88.62            & 46.14 \\ 
                                         MLS \tiny{(ICML'22)} & 98.57               & 6.53                & 89.24             & 46.65            & 91.50            &  38.97          & 93.05               &  30.27            &    93.09          & 30.61 \\
                                         Energy \tiny{(NeurIPS'20)} & 98.66               &  6.04              &  89.30             &  \underline{45.27}           & 91.75           &  37.09          &  93.42             & 28.12              &  93.28            & 29.13 \\
                                         ReAct \tiny{(NeurIPS'21)} & 99.01               & 4.19              & 89.11             & 47.37            & 91.54          & 39.04            & 93.38               & 28.37             & 93.26             & 29.74\\
                                         Maha \tiny{(NeurIPS'18)} & \textbf{99.63}               & \underline{2.15}               & 86.29             & 60.04            & 89.11           & 51.29           & \underline{94.21}               & \underline{25.27}            & 92.31             & 34.69 \\
                                         KL \tiny{(ICML'22)} & 96.62               & 16.45               & 84.34             & 63.50            & 85.91            & 59.41           & 88.26               & 45.99          & 88.78             & 46.34 \\
                                         ViM \tiny{(CVPR'22)} & 99.41                & 2.56               & 88.47             &  50.44            & 91.53            & 39.48            & \underline{95.31}              & \underline{20.14}              &  93.68            & 28.15 \\
                                         DICE \tiny{(ECCV'22)} & 65.49              & 91.69               & 61.56              & 93.88            & 64.75           & 88.51           & 75.49              & 74.26             &  66.82            & 87.09 \\
                                         ASH-S \tiny{(ICLR'23)} & 61.96               &  95.34              & 61.23             &  95.68           & 65.52           & 92.90           & 56.79              & 91.12              & 61.38             & 93.76 \\
                                         SHE \tiny{(ICLR'23)} & 90.69               & 43.50               &         85.63     & 59.25            & 87.97           &  51.93          & 92.18              & 32.31             & 89.12             & 46.75\\
                                         GEN \tiny{(CVPR'23)} & 98.63               & 5.83               & 89.22              &  49.61           & 91.00           & 43.36           & 92.35              & 34.05             & 92.80             & 33.21 \\
                                         PCA \tiny{(ICCV'23)} & 95.36              & 27.37                & 76.36              & 79.58            & 80.92          & 72.92            & 90.20              & 43.59             & 85.71             & 55.86 \\
                                         NNGuide \tiny{(ICCV'23)} & 87.25               & 52.54               & 80.85            & 62.61             & 86.04            & 51.16           & 91.97              & 32.66             &  88.03            & 46.05 \\
                                         SCALE \tiny{(ICLR'24)} & 98.88                & 5.00               & \underline{90.14}             & \textbf{42.89}            & \underline{92.57}           & \textbf{34.28}           & 93.39             & 28.51               & \underline{93.75}             & \underline{27.67} \\
                                         VRA \tiny{(NeurIPS'24)} & \underline{99.52}                &  \textbf{1.89}              &  \textbf{90.75}            & \underline{42.98}            & \textbf{92.99}        &  \underline{34.49}          & 93.37              & 29.19              & \underline{94.16}              & \underline{27.14} \\
                                         fDBD \tiny{(ICML'24)} & 98.39               & 7.15               & 89.07             &  48.68             &  91.13           & 41.35           & 93.10              & 29.81             & 92.92             & 31.75\\ \midrule
                                         $\text{NME}^+$   & 98.13       & 7.92  & 87.73        & 52.09       & 89.94      & 45.49      & 93.47         & 28.51        & 92.32        & 33.50        \\
       $\text{NME}^+$ + $\text{CO}^+$  & 98.59       & 6.22  & 88.18        & 50.98       & 90.22      & 44.57      & 93.78         & 27.19        & 92.69        & 32.24        \\
                                         MME & \underline{99.48}               & \underline{2.16}                & \underline{89.66}             & 46.71            & \underline{92.49}          & \underline{35.92}             & \textbf{95.48}              & \textbf{19.77}             & \textbf{94.28}             & \textbf{26.14} \\
                                         \bottomrule
\end{tabular}
\caption{Ablation study on impact of $\text{NME}^+$ and $\text{CO}^+$. Results on ImageNet-1K with ViT model. The best method is highlighted in bold, and the 2nd and 3rd best are underlined.}
\label{tab:nme}
\end{table*}

\subsection{CO performance with various scoring function}
\label{appendix:co}

In \cref{subsec:preli} and \cref{tab:co}, we validate Assumption \ref{ass:cov} using the CO function (\Cref{eq:co}). The CO function is designed for OOD detectors that also provide classification predictions, enabling direct comparison between predictions. Extending CO to modern scoring functions is non-trivial, as these typically do not produce classification outputs. Therefore, to further validate Assumption \ref{ass:cov}, we measure the covariance between six scoring functions on the ID dataset, as well as the average covariance across four OOD datasets. As shown in \cref{tab:co2}, the covariance values for the ID dataset are consistently higher than those for the OOD datasets, which further supports Assumption \ref{ass:cov} and \cref{prop:scoring}.

\begin{table}[]
\scriptsize
\begin{tabular}{c|ccccccc}
\toprule
ID / OOD & MLS & Energy & ViM & GEN & NNGuide & fDBD \\  \midrule
MLS &  & 4.44 / 3.88 & 6.12 / 5.26 & 16.03 / 15.47 & 9.00 / 8.21 & 109.76 / 105.19 \\
Energy & 4.44 / 3.88 &  & 5.62 / 4.16 & 14.48 / 11.80 & 8.29 / 6.48 & 99.09 / 78.81 \\
ViM & 6.12 / 5.26 & 5.62 / 4.16 &  & 20.18 / 15.81 & 11.15 / 8.75 & 145.68 / 110.26 \\
GEN & 16.03 / 15.47 & 14.48 / 11.80 & 20.18 / 15.81 &  & 26.72 / 23.39 & 370.00 / 319.26 \\
NNGuide & 9.00 / 8.21 & 8.29 / 6.48 & 11.15 / 8.75 & 26.72 / 23.39 &  & 199.56 / 162.68 \\
fDBD & 109.76 / 105.19 & 99.09 / 78.81 & 145.68 / 110.26 & 370.00 / 319.26 & 199.56 / 162.68 & 
\end{tabular}
\caption{Covariance between six state-of-the-art OOD detectors on the ID dataset ImageNet-1K / averaged across four OOD datasets (iNaturalist, Places, SUN, and Textures), using a ViT model.}
\label{tab:co2}
\end{table}

\section{Ablation study}
\label{appendix:ablation}

\begin{table}[]
\tiny
\begin{tabular}{cccccccccccc}
\toprule
                         \multirow{2}{*}{Method} &   \multirow{2}{*}{Feature truncation}     & \multicolumn{2}{c}{iNaturalist}       & \multicolumn{2}{c}{Places} & \multicolumn{2}{c}{SUN} & \multicolumn{2}{c}{Textures} & \multicolumn{2}{c}{Average} \\
                        &       & \tiny{AUROC$\uparrow$}       & \tiny{FPR95$\downarrow$} & \tiny{AUROC$\uparrow$}        & \tiny{FPR95$\downarrow$}       & \tiny{AUROC$\uparrow$}      & \tiny{FPR95$\downarrow$}      & \tiny{AUROC$\uparrow$}         & \tiny{FPR95$\downarrow$}        & \tiny{AUROC$\uparrow$}        & \tiny{FPR95$\downarrow$}        \\ \cmidrule(l){1-2} \cmidrule(l){3-10} \cmidrule(l){11-12}
\multirow{2}{*}{Energy} & Base  & 99.40       & 2.48  & 88.83        & 48.91       & 91.70      & 38.82      & 95.35         & 20.02        & 93.82        & 27.56        \\
                        & VRA   & \textbf{99.63}       & \textbf{1.57}  & 89.34        & 49.61       & 92.12      & 38.46      & 95.18         & 20.81        & 94.07        & 27.61        \\ \cmidrule(l){1-2} \cmidrule(l){3-10} \cmidrule(l){11-12}
\multirow{2}{*}{ViM}    & Base  & 99.51       & 2.04  & 89.55        & 46.94       & 92.46      & \textbf{35.82}      & \textbf{95.51}         & \textbf{19.65}        & 94.26        & 26.11        \\
                        & SCALE & 99.33       & 2.68  & 88.43        & 51.41       & 91.29      & 41.59      & 95.25         & 20.85        & 93.58        & 29.13        \\ \cmidrule(l){1-2} \cmidrule(l){3-10} \cmidrule(l){11-12}
\multirow{2}{*}{fDBD}   & Base  & 99.45       & 2.26  & 89.61        & 46.81       & 92.45      & 35.94      & 95.49         & 19.79        & 94.25        & 26.20        \\
                        & SCALE & 99.45       & 2.26  & 89.61        & 46.80       & 92.45      & 35.92      & 95.49         & 19.77        & 94.25        & 26.19        \\ \cmidrule(l){1-2} \cmidrule(l){3-10} \cmidrule(l){11-12}
\multirow{2}{*}{PCA}    & Base  & 99.46       & 2.25  & 89.61        & 46.74       & 92.44      & 35.98      & 95.47         & 19.83        & 94.24        & 26.20        \\
                        & SCALE & 99.46       & 2.25  & 89.61        & 46.74       & 92.44      & 35.98      & 95.47         & 19.83        & 94.24        & 26.20        \\ \cmidrule(l){1-2} \cmidrule(l){3-10} \cmidrule(l){11-12}
\multirow{2}{*}{$\text{NME}^+$}    & SCALE & 99.48       & 2.17  & 89.66        & 46.72       & 92.49      & 35.93      & 95.48         & 19.78        & 94.28        & 26.15        \\
                        & VRA   & 99.52       & 2.05  & \textbf{89.73}        & 46.77       & \textbf{92.54}      & 35.84      & 95.45         & 20.04        & 94.31        & 26.17        \\ \cmidrule(l){1-2} \cmidrule(l){3-10} \cmidrule(l){11-12}
\multicolumn{2}{c}{Baseline}    & 99.48       & 2.16  & 89.66        & \textbf{46.71}       & 92.49      & 35.92      & 95.48         & 19.77        & \textbf{94.28}        & \textbf{26.14}       \\
\bottomrule
\end{tabular}
\caption{Ablation study on the variation of feature truncation methods impact each component of MME. Results on ImageNet-1K with ViT model.}
\label{tab:abl1}
\end{table}

\subsection{Varying feature truncation} \cref{tab:abl1} illustrates how MME performance changes when different combinations of feature truncation and scoring functions are applied. Overall, we observe that the highest average performance is achieved with the default parameters. However, performance remains consistently stable, demonstrating the robustness of our ensemble approach.

\begin{table}[]
\tiny
\begin{tabular}{ccccccccccc}
\toprule
\multirow{2}{*}{Scoring function} & \multicolumn{2}{c}{iNaturalist}       & \multicolumn{2}{c}{Places} & \multicolumn{2}{c}{SUN} & \multicolumn{2}{c}{Textures} & \multicolumn{2}{c}{Average} \\
                               & AUROC$\uparrow$       & FPR95$\downarrow$ & AUROC$\uparrow$        & FPR95$\downarrow$       & AUROC$\uparrow$      & FPR95$\downarrow$      & AUROC$\uparrow$         & FPR95$\downarrow$        & AUROC$\uparrow$        & FPR95$\downarrow$               \\ \cmidrule(l){1-1} \cmidrule(l){2-9} \cmidrule(l){10-11}
GEN               & 99.15       & 3.77  & \textbf{89.84}        & 47.70       & 91.89      & 39.39      & 93.79         & 28.24        & 93.67        & 29.77        \\
NNGuide           & 99.43       & 2.36  & 89.49        & 47.15       & 92.41      & 36.10      & \textbf{95.51}         & \textbf{19.50}        & 94.21        & 26.28        \\
SHE               & 99.43       & 2.31  & 89.71        & 46.72       & 92.47      & 35.92      & 95.48         & 19.63        & 94.27        & 26.15        \\
$\varnothing$                & \textbf{99.48}       & \textbf{2.16}  & 89.66        & \textbf{46.71}       & \textbf{92.49}      & \textbf{35.92}      & 95.48         & 19.77        & \textbf{94.28 }       & \textbf{26.14}  \\ \bottomrule     
\end{tabular}
\caption{Ablation study on adding more scoring function to MME. Results on ImageNet-1K with ViT model.}
\label{tab:abl2}
\end{table}

\subsection{Adding more scoring function} \cref{tab:abl2} shows the impact of adding more scoring functions to MME. Empirically, we observe that adding more functions no longer improves performance. As explained by \cref{prop:scoring}, adding more functions does not necessarily enhance performance. In fact, adding more functions without novel information could compromise Assumption \ref{ass:cov}, leading to a degradation in performance.

\subsection{Effect of hyperparameter} \label{ssec:hyperpara}

We empirically analyze how the performance of MME varies with different values of $\lambda$ and $T$ in terms of AUROC.

\begin{figure}[t]
  \centering
   \includegraphics[width=0.9\linewidth]{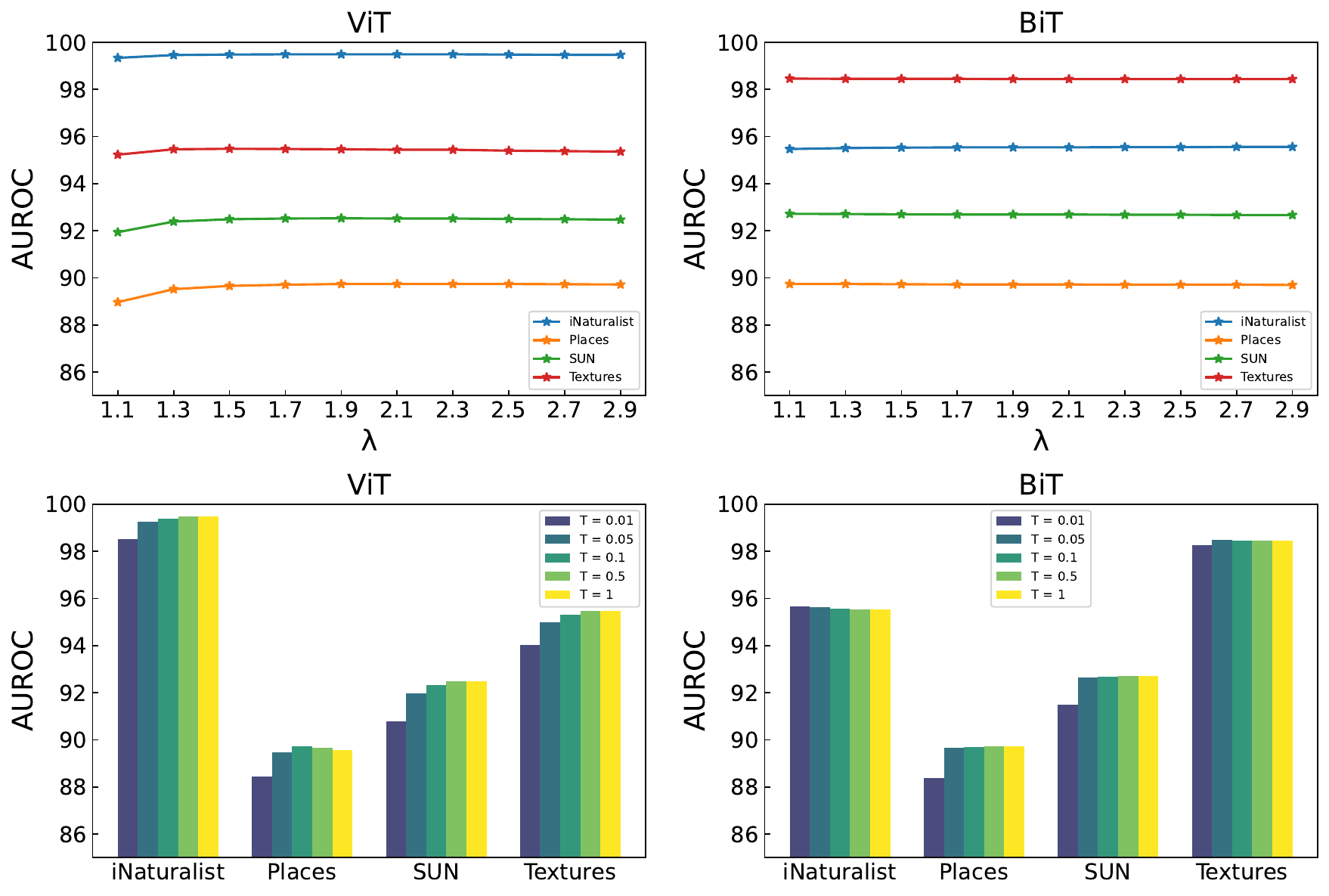}

   \caption{Ablation study on the impact of $\lambda$ and $T$ hyperparameters. The AUROC performance of MME is reported on the ImageNet-1K benchmark for different values of $\lambda$ (top row) and $T$ (bottom row). The results are shown for the ViT architecture on the left and the BiT architecture on the right.}
   \label{fig:hyperpara}
\end{figure}

First, we examine the effect of $\lambda$. The first row of \Cref{fig:hyperpara} shows that the best performance is achieved with $\lambda=1.5$ for both ViT \cite{dosovitskiy2020image} and BiT \cite{kolesnikov2020big} architectures. However, once this value increases further, the performance remains unchanged, indicating that MME is relatively robust to the choice of $\lambda$.

We also investigate the effect of different values of $T$. The second row of \Cref{fig:hyperpara} demonstrates that very low values of $T$ ($T<0.1$) lead to suboptimal performance. Performance improves as $T$ increases, reaching its peak at $T=0.5$. Beyond this point, further increases in $T$ result in a performance plateau or a slight decline, suggesting that excessive smoothing of predictions is unnecessary.

The current evaluation protocol for OOD detection relies on the test dataset, which may not be ideal for real-world applications. However, as shown in \Cref{fig:hyperpara}, MME remains highly robust to variations in hyperparameters due to its ensemble strategy. Unless $\lambda$ or $T$ undergoes extreme changes, performance remains stable, with less than a 2\% difference between the best and worst cases. Additionally, MME was evaluated on completely unseen datasets, including NINCO \cite{bitterwolf2023or} and SSB-hard \cite{vaze2021open} for near-OOD detection and OpenImage-O \cite{wang2022vim} for far-OOD detection and it consistently achieves state-of-the-art performance, as illustrated in \Cref{fig:near}.

\section{Dataset description}
\label{appendix:dataset}

\textbf{Large-scale.} For large-scale evaluation, we use ImageNet-1k \cite{russakovsky2015imagenet} as the in-distribution dataset and evaluate on four OOD datasets following the setup proposed by

\begin{itemize}
\item \textbf{iNaturalist} \cite{van2018inaturalist} dataset contains 859,000 images of plants and animals, covering over 5,000 different species. Each image is resized to have a maximum dimension of 800 pixels. For evaluation, we use a random sample of 10,000 images from 110 classes that do not overlap with those in ImageNet-1k.
\item \textbf{Places} \cite{zhou2017places} dataset is a scene dataset. For evaluation, we use a selected subset of 10,000 images spanning 50 classes that are not included in ImageNet-1k.
\item \textbf{SUN} \cite{xiao2010sun} dataset comprises over 130,000 images of scenes across 397 categories. Although there are overlapping categories between SUN and ImageNet-1k, our evaluation uses 10,000 images randomly sampled from 50 classes that do not overlap with ImageNet-1k labels.
\item \textbf{Textures} \cite{cimpoi2014describing} dataset includes 5,640 real-world texture images categorized into 47 different classes. We utilize the entire dataset for our evaluation.
\end{itemize}

\textbf{Small-scale.} The CIFAR \cite{krizhevsky2009learning} benchmarks, CIFAR-10 and CIFAR-100, are widely recognized ID datasets in the literature, containing 10 and 100 classes, respectively. We follow the standard split, using 50,000 training images and 10,000 test images. Our approach is evaluated using six common OOD datasets, which are listed below:

\begin{itemize}
    \item \textbf{SVHN} \cite{netzer2011reading} consists of color images of house numbers, categorized into ten digit classes (0-9). We use the entire test set, which includes 26,032 images.
    \item \textbf{LSUN-Crop} \cite{yu2015lsun} comprises 10,000 testing images across 10 different scenes. For evaluation, image patches of size 32$\times$32 are randomly cropped from this dataset.
    \item \textbf{LSUN-Resize} \cite{yu2015lsun} comprises 10,000 testing images across 10 different scenes. For evaluation, image patches of size 32$\times$32 are randomly resized from this dataset.
    \item \textbf{iSUN} \cite{xu2015turkergaze} provides ground truth gaze traces on images from the SUN dataset. Following prior work, we use a subset of 8,925 images for our evaluation.
    \item \textbf{Places365} \cite{zhou2017places} is a large-scale dataset including photographs of scenes categorized into 365 scene types, with 900 images per category in the test set, totaling 328,500 images. We utilize the entire test set for our evaluation.
    \item \textbf{Textures} \cite{cimpoi2014describing} dataset includes 5,640 real-world texture images categorized into 47 different classes. We utilize the entire dataset for our evaluation.
\end{itemize}

\textbf{Near and far-OOD.} Recent works \cite{xu2023scaling, zhang2023openood} differentiate between near-OOD and far-OOD cases. To remain consistent with these studies, we adopt their settings: ImageNet-1K is used as the ID set while the following datasets are used for near-OOD evaluation:

\begin{itemize}
    \item \textbf{SSB-hard} \cite{vaze2021open} is the hard split of the Semantic Shift Benchmark (SSB), which is more aligned with detecting semantic novelty as opposed to other distributional shifts addressed by related fields such as out-of-distribution detection. We use the entire test set comprising 49,000 images.
    \item \textbf{NINCO} \cite{bitterwolf2023or} is a dataset containing images that do not include objects from any of the 1000 classes of ImageNet-1K, making it suitable for evaluating out-of-distribution detection on ImageNet-1K. The NINCO dataset consists of 64 OOD classes with a total of 5,879 samples. Each sample was individually inspected by the authors to ensure it does not contain ID objects.
\end{itemize}

The following datasets are used for far-OOD evaluation:

\begin{itemize}
    \item \textbf{iNaturalist} \cite{van2018inaturalist} dataset contains 859,000 images of plants and animals, covering over 5,000 different species. Each image is resized to have a maximum dimension of 800 pixels. For evaluation, we use a random sample of 10,000 images from 110 classes that do not overlap with those in ImageNet-1k.
    \item \textbf{Textures} \cite{cimpoi2014describing} dataset includes 5,640 real-world texture images categorized into 47 different classes. We utilize the entire dataset for our evaluation.
    \item \textbf{OpenImage-O} \cite{wang2022vim} dataset is curated image-by-image from the test set of OpenImage-V3, collected from Flickr. It does not have a predefined list of class names or tags, which results in natural class statistics and avoids initial design bias.
\end{itemize}
\end{document}